\documentclass[conference]{IEEEtran}

\usepackage[sorting=none]{biblatex} 
\AtNextBibliography{\small}
\ifCLASSINFOpdf
  \usepackage[pdftex]{graphicx}
  \graphicspath{{./figures}}
\else
\fi
\usepackage{amsmath}
\usepackage{amssymb}
\usepackage{algorithm}
\usepackage{algpseudocode}

\usepackage{array}
\makeatletter
 \let\old@ps@headings\ps@headings
 \let\old@ps@IEEEtitlepagestyle\ps@IEEEtitlepagestyle
 \def\confheader#1{%
 \def\ps@headings{%
 \old@ps@headings%
 \def\@oddhead{\strut\hfill#1\hfill\strut}%
 \def\@evenhead{\strut\hfill#1\hfill\strut}%
 }%
 \def\ps@IEEEtitlepagestyle{%
 \old@ps@IEEEtitlepagestyle%
 \def\@oddhead{\strut\hfill#1\hfill\strut}%
 \def\@evenhead{\strut\hfill#1\hfill\strut}%
 }%
 \ps@headings%
 }
 \makeatother

\confheader{%
2nd International Conference on Image Processing, Computer Vision and Machine Learning, November 2023
 }

\begin{document}
%
\title{Novel View Synthesis from a Single RGBD Image for Indoor Scenes}

\author{\IEEEauthorblockN{Congrui Hetang*}
\IEEEauthorblockA{
\textit{Carnegie Mellon University}\\
Pittsburgh, PA 15213 \\
congruihetang@gmail.com}
\and
\IEEEauthorblockN{Yuping Wang}
\IEEEauthorblockA{
\textit{University of Michigan}\\
Ann Arbor, MI 48109 \\
ypw@umich.edu}}


%


\maketitle

\begin{abstract}
In this paper, we propose an approach for synthesizing novel view images from a single RGBD (Red Green Blue-Depth) input. Novel view synthesis (NVS) is an interesting computer vision task with extensive applications. Methods using multiple images has been well-studied, exemplary ones include training scene-specific Neural Radiance Fields (NeRF), or leveraging multi-view stereo (MVS) and 3D rendering pipelines. However, both are either computationally intensive or non-generalizable across different scenes, limiting their practical value. Conversely, the depth information embedded in RGBD images unlocks 3D potential from a singular view, simplifying NVS. The widespread availability of compact, affordable stereo cameras, and even LiDARs in contemporary devices like smartphones, makes capturing RGBD images more accessible than ever. In our method, we convert an RGBD image into a point cloud and render it from a different viewpoint, then formulate the NVS task into an image translation problem. We leveraged generative adversarial networks to style-transfer the rendered image, achieving a result similar to a photograph taken from the new perspective. We explore both unsupervised learning using CycleGAN and supervised learning with Pix2Pix, and demonstrate the qualitative results. Our method circumvents the limitations of traditional multi-image techniques, holding significant promise for practical, real-time applications in NVS.
\end{abstract}


%
\IEEEpeerreviewmaketitle

\section{Introduction}
The capability to synthesize accurate and realistic novel views of a scene is under active research by the computer vision community, and it has profound application value in virtual or augmented reality \cite{mihelj2014virtual}, scene understanding \cite{rombach2021geometry,congrui_region, zhang2021point} and editing, simulation for robotics and autonomous vehicles \cite{ku2019improving, wang2023equivariant}, and more. Traditionally, creating these novel viewpoints has required either a series of multi-view images, and intricate 3D modeling techniques. However, these methods are computationally intensive, making them unsuitable for real-time applications \cite{congrui_depth}, and lacks flexibility to generalize across different scenes. The emergence of RGBD sensors, which capture both color and depth data, makes it possible to generate novel views from a single image, and opens up an avenue for a much simpler and streamlined single-shot NVS approach.

In this work, we propose a one-shot novel view synthesis approach from a single RGBD image. We directly reproject the RGBD image into a new viewpoint with the given camera parameters, then use generative adversarial networks to fix the artifacts in the reprojected image to produce a plausible photorealistic novel-view image. We proved the feasibility of forming the NVS task into an image translation problem, and verified the effectiveness of the proposed method on the SUN3D dataset. 

\begin{figure*}[t]
\centering
\includegraphics[scale=0.65]{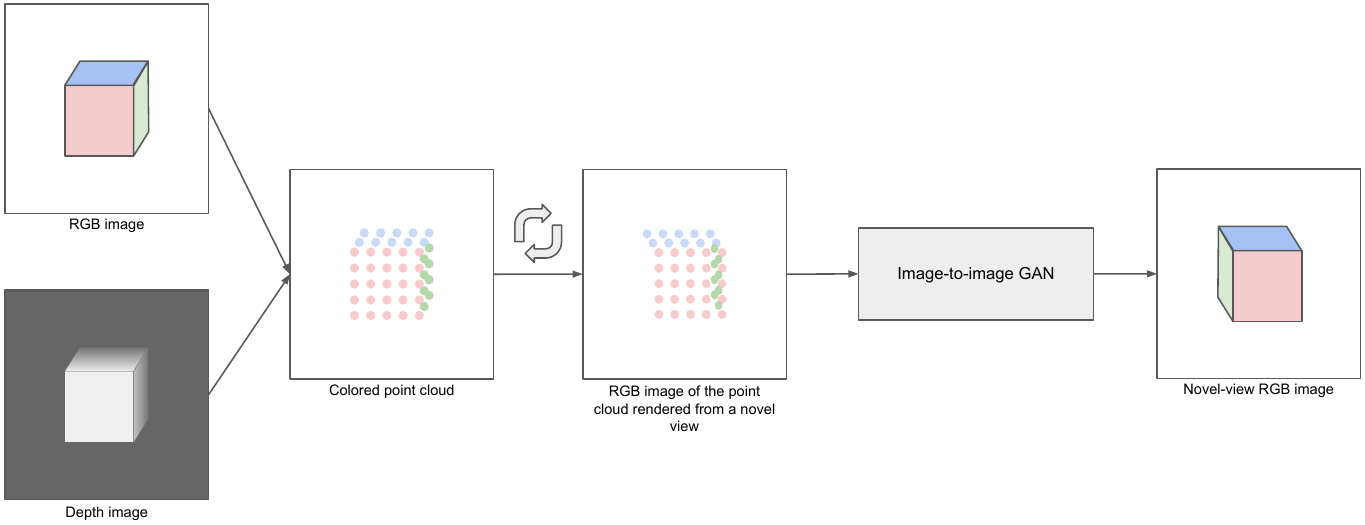}
\caption[position=bottom]{The overall pipeline of the proposed method.}
\label{fig:method}
\end{figure*}

\section{Related Works}

\subsection{Multi-view Image Generation}

Traditional methods for generating multi-view images rely on capturing multiple photographs from different angles, as discussed in works like \cite{sun2018multi}. These approaches can be both time-consuming and computationally costly, making automated solutions increasingly attractive. Recent works, such as \cite{zhou2018stereo}, have started to employ machine learning techniques to automate this process, albeit still requiring multiple images for training.

\subsection{Single Image Novel View Synthesis}
Novel View Synthesis (NVS) from a single image has been a subject of considerable interest. Techniques have varied from geometric transformations \cite{schonberger2016structure} to deep learning models like GANs \cite{goodfellow2014generative}. \cite{wu2016single} pioneered unsupervised learning techniques for this purpose, inspiring our initial approach.

\subsection{RGBD Image Utilization}
The inclusion of depth information in image data (RGBD) has shown significant promise in various computer vision tasks. While depth information has been traditionally used for object recognition and segmentation \cite{silberman2012indoor}, recent research has started exploring its utility in NVS \cite{zhu2018generative}.

\subsection{Image-to-Image Translation}
Image-to-image translation models like Cycle-GAN \cite{zhu2017unpaired} and pix2pix \cite{isola2017image} have demonstrated proficiency in tasks that involve changing the content or style of images. These models have recently been adapted for multi-view image generation \cite{zhao2018multi}.

\subsection{Supervised Methods in NVS}
Supervised approaches have also garnered attention, especially with the availability of datasets that provide ground truth data. ScanNet \cite{dai2017scannet}, which provides video frames annotated with extrinsic parameters, has enabled the development of supervised methods for NVS. This dataset plays a pivotal role in our supervised approach.

\subsection{Neural Radiance Fields}
Neural Radiance Fields (NeRF) models, initially introduced by  \cite{mildenhall2021nerf}, have gained substantial attention for their ability to synthesize novel 3D views from a sparse set of 2D images. NeRF models operate by learning a continuous volumetric scene representation, capturing both color and spatial variances, making them highly effective in rendering photorealistic novel views of a 3D scene. However, these models come with significant drawbacks. First and foremost, they are computationally intensive. The NeRF architecture necessitates a large number of network evaluations for rendering each new view, leading to high computational costs. This makes real-time application or deployment on resource-constrained devices impractical. Secondly, NeRF models are often criticized for their lack of generalization. While they excel in learning a detailed 3D representation from a specific dataset, their ability to generalize to unseen data or different scenes is limited. This implies that a separate model would have to be trained for each new scene or set of conditions, further exacerbating the computational burden. Lastly, although NeRF models offer high-quality outputs, they usually require substantial amounts of training data and extensive hyperparameter tuning to achieve optimal performance, making them less flexible for broader applications.

Given these limitations, our research aims to explore alternative methodologies that are both computationally efficient and more generalizable, particularly for the challenging settings of indoor scenes.

\section{Proposed Method}

The pipeline of our proposed method is shown in Figure ~\ref{fig:method}. An RGBD image contains per-pixel depth information. By utilizing the camera's intrinsics and extrinsics, we can map each pixel to a corresponding point in 3D space, effectively transforming the entire RGBD image into a 3D point cloud. Each point in this cloud possesses (x, y, z) coordinates along with RGB color information.

Our objective is to generate an RGB image of the scene from a novel viewpoint. This goal necessitates a new set of camera intrinsics and extrinsics. With these parameters, we can re-project the aforementioned point cloud onto the image plane using the pinhole camera model. This process involves rendering each point to a pixel and assigning the appropriate color, resulting in an RGB image akin to what one would observe when manipulating a point cloud in 3D visualization tools. We refer to this image as the "re-projected image."

Although the re-projected image maintains structural correctness due to the depth information, it suffers from obvious artifacts. These deficiencies, such as missing areas, result from occlusions, depth inaccuracies, and point sparsity. To acquire a realistic novel-view image, we need to fix these artifacts. Consequently, we approach this challenge as a combination of image inpainting and style transfer. We employ a Generative Adversarial Network (GAN) to perform style transfer, aligning the re-projected image with the appearance of actual photographs. This technique yields the final synthetic novel-view RGB image.

We experimented two methods for image translation: unsupervised (no paird images) and supervised (with paired images).

\subsection{Unsupervised Method}
\begin{figure}[ht]
\centering
\includegraphics[scale=0.6]{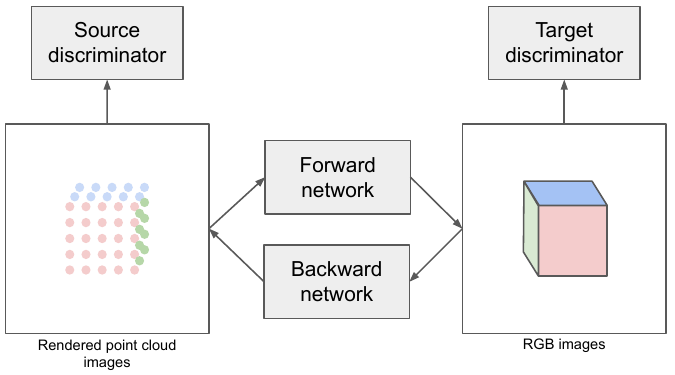}
\caption[position=bottom]{Unpaired image translation with CycleGAN. Notice the source and target domain images are not paired, in our case they don't need to have the same viewpoint.}
\label{fig:cycle}
\end{figure}

For the unpaired training experiments, we utilized CycleGAN, aiming to refine the projected point cloud image so it resembles an authentic photograph. The training process is showin in Figure ~\ref{fig:cycle}. We generated training images from the SUN 3D dataset, varying camera positions randomly. The reprojected point cloud images forms the source domain, and the real RGB images forms the target domain.

\subsection{Supervised Method}
\begin{figure}[ht]
\centering
\includegraphics[scale=0.6]{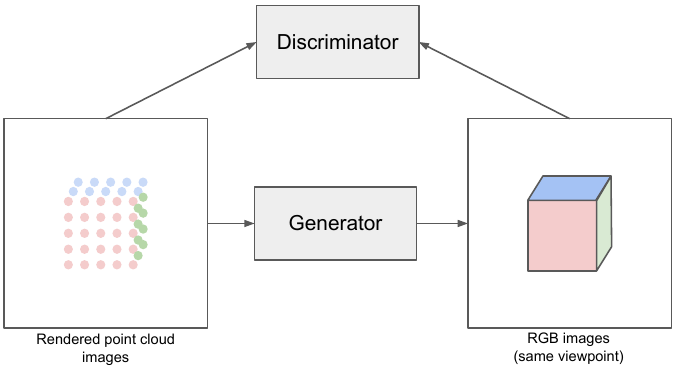}
\caption[position=bottom]{Paired image translation with Pix2Pix. Here the source and target domain images are paired, which means they have the same viewpoint.}
\label{fig:pix2pix}
\end{figure}

We extracted paired images from RGBD videos of indoor scenes. They comprises numerous RGBD frames with precise camera parameters. Our approach selects two proximate frames, A and B; we projected the point cloud from view A into view B and used the RGB image of view B as the ground truth. This procedure enabled the generation of thousands of instances of paired training data, thereby facilitating the utilization of the Pix2Pix method for paired image translation. The training process is shown by Figure ~\ref{fig:pix2pix}.


\section{Experiment Details}

\begin{figure}[ht]
\centering
\includegraphics[width=0.2\textwidth]{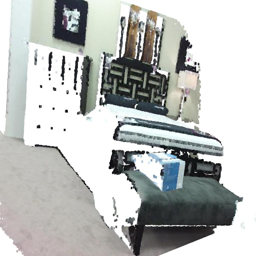}%
\hspace{0.02\textwidth}%
\includegraphics[width=0.2\textwidth]{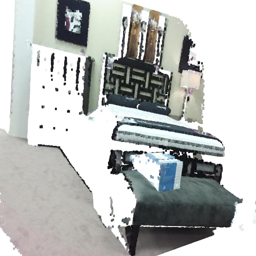}

\vspace{0.02\textwidth} 

\includegraphics[width=0.2\textwidth]{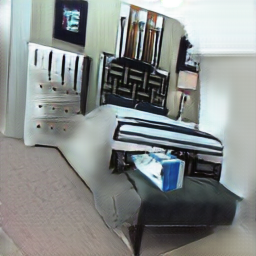}%
\hspace{0.02\textwidth}%
\includegraphics[width=0.2\textwidth]{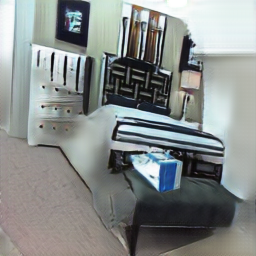}
\caption{CycleGAN results on SUN3D datasets. The top row shows the original projected images, while the bottom row displays our generated images.}
\label{fig:cyc}
\end{figure}

\begin{figure}[ht]
\centering
\includegraphics[width=0.2\textwidth]{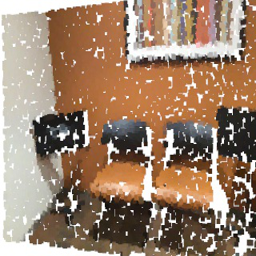}%
\hspace{0.02\textwidth}%
\includegraphics[width=0.2\textwidth]{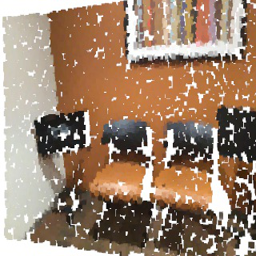}

\vspace{0.02\textwidth} 

\includegraphics[width=0.2\textwidth]{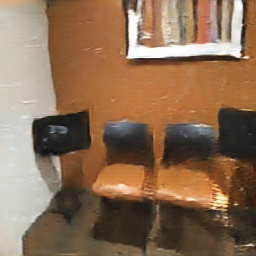}%
\hspace{0.02\textwidth}%
\includegraphics[width=0.2\textwidth]{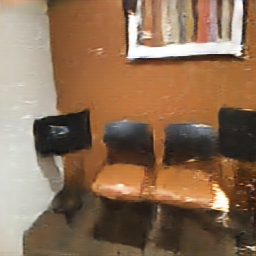}
\caption{Pixel2Pixel results. Top: Original projected image. Bottom: Our generated images.}
\label{fig:pix}
\end{figure}

\subsection{Data Generation}

We employed SUN3D to generate our training datasets. The SUN3D dataset is a comprehensive resource for 3D scene understanding research. It contains a diverse collection of RGB-D video sequences captured from a variety of indoor scenes. The sequences contain complete camera intrinsics and extrinsics parameters.

For the unsupervised setup, we generate a non-paired training dataset. This means the source image and target image don't need to have aligned contents, in our case, the same viewpoint. Utilizing Open3D, we applied specific rotation angles to the original images, creating the new viewpoints. As expected, these reprojected images inherently possessed noise and inconsistencies. To form the source domain for image translation, we created 30,000 such images. Their original version forms the target domain. There are no image to image pairing between the source and target domains. This dataset allow us to train a CycleGAN, capable of unpaired image style transfer. 

For the supervised setup, we generate a paired training dataset. For this setup, each reprojected image is paired with a real image, captured from the identical viewpoint. This is achieved by leveraging the RGBD sequences in the SUN3D dataset - they naturally provide varying camera parameters along the time axis. Given one "original" frame, we randomly find a "novel" frame within the next 10 frames (this embodies the idea of \cite{congrui_impression}); the camera pose of the new frame serves as the "novel view". The reprojected point cloud image of the "original" frame and the RGB image of the "novel" frame forms a (reprojected image, real image) pair, where they have identical viewpoints. We selected 40 sequences, each contributing 50 pairs. This dataset allows us to train a Pix2Pix for paired image style transfer.

\subsection{Training Details}
We utilized the official CycleGAN and Pix2Pix codebases, tailoring them to our project's needs. We used the standard U-Net architecture for our generator networks. Following the standard practice ~\cite{congrui_unlock,congrui_sliding}, we fine-tune from pre-trained model weights. Our experiments were conducted on V100 GPUs. The training for both methods continued until the models began generating visually convincing images. 

\section{Results}
The qualitative results of the unsupervised CycleGAN methodology are depicted in Fig. \ref{fig:cyc}. The network succeeds in refining surfaces and filling in the voids with plausible colors. The qualitative results of the supervised Pix2Pix approach is showcased in Fig. \ref{fig:pix}. Intricate object details, such as the vertical sections of desks or the sides of chairs, are recovered. Rather than merely blurring out imperfections, the network discerns and replicates local patterns consistent with the scene's 3D structure.

For quantitative metrics, we conducted a user study to compare human preferences over different methods to synthesize the novel view, given the same reprojected point cloud image. In addition to CycleGAN and Pix2Pix, we trained a simple image completion network on the paired dataset just with the MSE loss, without the GAN training procedures. 12 volunteers were presented 10 sets of synthetic images each, and were asked to select the most visually appealing one. The frequencies of the MSE baseline, CycleGAN and Pix2Pix results being selected as the best are 10.8\%, 26.7\% and 62.5\%, respectively. We therefore conclude that using paired data with generative training goals yields the best performance.

\section{Conclusion}

To conclude, our work validated the concept of performing novel view synthesis using a single RGBD image, approached as an image translation task. Our method's one-shot nature during inference and its ability to generalize across different scenes once trained indicates its practical value. Future work could consider integrating advanced techniques like diffusion models for improved image generation, to achieve even better results.

\printbibliography

@inproceedings{zhao2018multi,
  title={Multi-view image generation from a single-view},
  author={Zhao, Bo and Wu, Xiao and Cheng, Zhi-Qi and Liu, Hao and Jie, Zequn and Feng, Jiashi},
  booktitle={2018 ACM Multimedia Conference on Multimedia Conference},
  pages={383--391},
  year={2018},
  organization={ACM}
}

@inproceedings{wu2016single,
  title={Single image 3d interpreter network},
  author={Wu, Jiajun and Xue, Tianfan and Lim, Joseph J and Tian, Yuandong and Tenenbaum, Joshua B and Torralba, Antonio and Freeman, William T},
  booktitle={European Conference on Computer Vision},
  pages={365--382},
  year={2016},
  organization={Springer}
}

@inproceedings{zhu2018generative,
  title={Generative Adversarial Frontal View to Bird View Synthesis},
  author={Zhu, Xinge and Yin, Zhichao and Shi, Jianping and Li, Hongsheng and Lin, Dahua},
  booktitle={2018 International Conference on 3D Vision (3DV)},
  pages={454--463},
  year={2018},
  organization={IEEE}
}

@inproceedings{zhu2017unpaired,
  title={Unpaired image-to-image translation using cycle-consistent adversarial networks},
  author={Zhu, Jun-Yan and Park, Taesung and Isola, Phillip and Efros, Alexei A},
  booktitle={Proceedings of the IEEE international conference on computer vision},
  pages={2223--2232},
  year={2017}
}

@inproceedings{isola2017image,
  title={Image-to-image translation with conditional adversarial networks},
  author={Isola, Phillip and Zhu, Jun-Yan and Zhou, Tinghui and Efros, Alexei A},
  booktitle={Proceedings of the IEEE conference on computer vision and pattern recognition},
  pages={1125--1134},
  year={2017}
}

@inproceedings{dai2017scannet,
  title={Scannet: Richly-annotated 3d reconstructions of indoor scenes},
  author={Dai, Angela and Chang, Angel X and Savva, Manolis and Halber, Maciej and Funkhouser, Thomas and Nie{\ss}ner, Matthias},
  booktitle={Proceedings of the IEEE Conference on Computer Vision and Pattern Recognition},
  pages={5828--5839},
  year={2017}
}

@inproceedings{sun2018multi,
  title={Multi-view to novel view: Synthesizing novel views with self-learned confidence},
  author={Sun, Shao-Hua and Huh, Minyoung and Liao, Yuan-Hong and Zhang, Ning and Lim, Joseph J},
  booktitle={Proceedings of the European Conference on Computer Vision (ECCV)},
  pages={155--171},
  year={2018}
}

@article{mildenhall2021nerf,
  title={Nerf: Representing scenes as neural radiance fields for view synthesis},
  author={Mildenhall, Ben and Srinivasan, Pratul P and Tancik, Matthew and Barron, Jonathan T and Ramamoorthi, Ravi and Ng, Ren},
  journal={Communications of the ACM},
  volume={65},
  number={1},
  pages={99--106},
  year={2021},
  publisher={ACM New York, NY, USA}
}

@inproceedings{silberman2012indoor,
  title={Indoor segmentation and support inference from rgbd images},
  author={Silberman, Nathan and Hoiem, Derek and Kohli, Pushmeet and Fergus, Rob},
  booktitle={Computer Vision--ECCV 2012: 12th European Conference on Computer Vision, Florence, Italy, October 7-13, 2012, Proceedings, Part V 12},
  pages={746--760},
  year={2012},
  organization={Springer}
}

@article{zhou2018stereo,
  title={Stereo magnification: Learning view synthesis using multiplane images},
  author={Zhou, Tinghui and Tucker, Richard and Flynn, John and Fyffe, Graham and Snavely, Noah},
  journal={arXiv preprint arXiv:1805.09817},
  year={2018}
}

@InProceedings{schonberger2016structure,
author = {Schonberger, Johannes L. and Frahm, Jan-Michael},
title = {Structure-From-Motion Revisited},
booktitle = {Proceedings of the IEEE Conference on Computer Vision and Pattern Recognition (CVPR)},
month = {June},
year = {2016}
}

@article{goodfellow2014generative,
  title={Generative adversarial nets},
  author={Goodfellow, Ian and Pouget-Abadie, Jean and Mirza, Mehdi and Xu, Bing and Warde-Farley, David and Ozair, Sherjil and Courville, Aaron and Bengio, Yoshua},
  journal={Advances in neural information processing systems},
  volume={27},
  year={2014}
}

@article{congrui_region, title={Region-Based Quality Estimation Network for Large-Scale Person Re-Identification}, volume={32}, url={https://ojs.aaai.org/index.php/AAAI/article/view/12305}, DOI={10.1609/aaai.v32i1.12305}, number={1}, journal={Proceedings of the AAAI Conference on Artificial Intelligence}, author={Song, Guanglu and Leng, Biao and Liu, Yu and Hetang, Congrui and Cai, Shaofan}, year={2018}, month={Apr.} }

@INPROCEEDINGS{congrui_impression,
  author={Hetang, Congrui},
  booktitle={2023 IEEE 3rd International Conference on Information Technology, Big Data and Artificial Intelligence (ICIBA)}, 
  title={Impression Network for Video Object Detection}, 
  year={2023},
  volume={3},
  number={},
  pages={735-743},
  doi={10.1109/ICIBA56860.2023.10165600}
}

@article{congrui_depth,
  title={Depth-wise decomposition for accelerating separable convolutions in efficient convolutional neural networks},
  author={He, Yihui and Qian, Jianing and Wang, Jianren},
  journal={arXiv preprint arXiv:1910.09455},
  year={2019}
}

@Article{congrui_unlock,
AUTHOR = {Zhou, Zhihao and Yue, Tianwei and Liang, Chen and Bai, Xiaoyu and Chen, Dachi and Hetang, Congrui and Wang, Wenping},
TITLE = {Unlocking Everyday Wisdom: Enhancing Machine Comprehension with Script Knowledge Integration},
JOURNAL = {Applied Sciences},
VOLUME = {13},
YEAR = {2023},
NUMBER = {16},
ARTICLE-NUMBER = {9461},
URL = {https://www.mdpi.com/2076-3417/13/16/9461},
ISSN = {2076-3417},
DOI = {10.3390/app13169461}
}

@article{congrui_sliding,
  title={Sliding-Bert: Striding Towards Conversational Machine Comprehension in Long Contex},
  author={Zhang, Longxiang and Wang, Wenping and Yu, Keyi and Huang, Jingxian and Lyu, Qi and Xue, Haoru and Hetang, Congrui},
  year={2023}
}

@article{mihelj2014virtual,
  title={Virtual reality technology and applications},
  author={Mihelj, Matja{\v{z}} and Novak, Domen and Begu{\v{s}}, Samo},
  year={2014},
  publisher={Springer}
}

@inproceedings{rombach2021geometry,
  title={Geometry-free view synthesis: Transformers and no 3d priors},
  author={Rombach, Robin and Esser, Patrick and Ommer, Bj{\"o}rn},
  booktitle={Proceedings of the IEEE/CVF International Conference on Computer Vision},
  pages={14356--14366},
  year={2021}
}

@inproceedings{ku2019improving,
  title={Improving 3d object detection for pedestrians with virtual multi-view synthesis orientation estimation},
  author={Ku, Jason and Pon, Alex D and Walsh, Sean and Waslander, Steven L},
  booktitle={2019 IEEE/RSJ International Conference on Intelligent Robots and Systems (IROS)},
  pages={3459--3466},
  year={2019},
  organization={IEEE}
}

@article{wang2023equivariant,
  title={Equivariant Map and Agent Geometry for Autonomous Driving Motion Prediction},
  author={Wang, Yuping and Chen, Jier},
  journal={arXiv preprint arXiv:2310.13922},
  year={2023}
}

@article{zhang2021point,
  title={Point set voting for partial point cloud analysis},
  author={Zhang, Junming and Chen, Weijia and Wang, Yuping and Vasudevan, Ram and Johnson-Roberson, Matthew},
  journal={IEEE Robotics and Automation Letters},
  volume={6},
  number={2},
  pages={596--603},
  year={2021},
  publisher={IEEE}
}

\end{document}